\documentclass{article} 
\PassOptionsToPackage{hyperfootnotes=false}{hyperref}
\PassOptionsToPackage{dvipsnames}{xcolor}
\usepackage{arxiv}

\usepackage{amsmath,amsfonts,bm}

\def\eqref#1{equation~\ref{#1}}

\def\1{\bm{1}}

\DeclareMathAlphabet{\mathsfit}{\encodingdefault}{\sfdefault}{m}{sl}
\SetMathAlphabet{\mathsfit}{bold}{\encodingdefault}{\sfdefault}{bx}{n}

\usepackage{longtable}

\usepackage{hyperref}
\usepackage{url}
\usepackage{xcolor}
\newcommand{\revision}[1]{#1}
\newcommand{\newrev}[1]{#1}
\usepackage{wrapfig}
\usepackage{subcaption}
\usepackage{cleveref}
\usepackage{enumitem}
\usepackage{amsmath}
\usepackage{graphicx}
\usepackage{makecell}
\usepackage{booktabs}
\usepackage{natbib}
\usepackage{array}
\usepackage[section]{placeins}
\usepackage{pifont}
\definecolor{accent}{RGB}{35,85,160}

\usepackage{xcolor}
\usepackage{colortbl}
\usepackage{pifont}
\usepackage{tcolorbox}
\tcbuselibrary{breakable,skins}
\usepackage{fvextra}
\usepackage[scaled=0.82]{beramono}

\definecolor{ourrow}{RGB}{252,239,226}
\definecolor{tracebg}{RGB}{247,248,250}
\definecolor{traceframe}{RGB}{194,201,207}

\newcommand{\mail}[1]{\href{mailto:#1}{\textcolor{black}{\texttt{#1}}}}
\newcommand{\cmark}{\textcolor{green!55!black}{\ding{51}}}
\newcommand{\xmark}{\textcolor{red!70!black}{\ding{55}}}

\setlist{noitemsep, topsep=2pt, parsep=0pt, partopsep=0pt}

\definecolor{accent}{RGB}{30,90,180}
\newcommand{\eqcontrib}{\textcolor{accent}{\scalebox{1.15}{\ding{79}}}}
\title{\revision{LeanPRM: Lean-Grounded Process Supervision for Natural Language Proof Verification}}
\date{}

\author{%
  \bfseries
  Rodney Lafuente-Mercado\,\textsuperscript{\eqcontrib\,1}\quad
  Salman Rahman\,\textsuperscript{\eqcontrib\,2,3}\quad
  Jingyan Shen\,\textsuperscript{3}\\[0.4em]
  \bfseries
  James Shiffer\,\textsuperscript{2}\quad
  Saadia Gabriel\,\textsuperscript{2}\quad
  Pavel Izmailov\,\textsuperscript{3}\\[0.7em]
  \normalfont\normalsize
  \textsuperscript{1}University of Texas at Austin\quad
  \textsuperscript{2}University of California, Los Angeles\quad
  \textsuperscript{3}New York University\\[0.5em]
  \small
  \eqcontrib\ Equal contribution
}

\definecolor{RoyalPurple}{RGB}{108, 59, 170}

\begin{document}

\maketitle
{\renewcommand{\thefootnote}{}%
 \footnotetext{Correspondence to: \mail{rlafuente@utexas.edu}, \mail{salman@nyu.edu}, \mail{pi390@nyu.edu}}}

\begin{abstract}
AI systems now write long and difficult mathematical proofs, and checking their step-by-step validity is becoming
a bottleneck. A learned verifier could check proofs cheaply, but training one requires a correctness
label on every step, and such labels are hard to obtain for open-ended proofs. Existing methods take
them from human annotators, which is costly, or derive them from final answers, which proofs do not
have, and formalizing each proof in Lean requires expert effort. We introduce LeanPRM, a framework
that uses the Lean proof assistant to create step-level labels for natural-language proofs. A model
writes a natural-language proof together with a paired formal proof, without feedback from Lean.
Lean then labels every step, and a second model discards failures caused only by the formal syntax,
so that the negative labels mark errors in the mathematics. LeanPRM contains 10{,}634 proofs with
152K labeled steps and no human annotation. A
Qwen3-4B verifier trained on LeanPRM raises exact step-label accuracy on a held-out set of
200 proofs (LeanPRM-Bench) from 48.2 to  \newrev{74.7}, as opposed to  \newrev{60.3 }when the same base model is trained on PRM800K. It also selects
better proofs in best-of-$N$ selection on IMO-level problems, and outperforms the PRM800K-trained verifier on three of four
external verification benchmarks. Training on LeanPRM improves verifiers at every size from 0.6B to 8B parameters on
LeanPRM-Bench, Hard2Verify and IMO-GradingBench.
Because LeanPRM requires no human annotation, it can scale with the capabilities of frontier provers, turning progress in proof generation into stronger verifier models.


\noindent Project page: \url{https://rlafuente.com/lean-prm}
\end{abstract}



\section{Introduction}

As AI systems become more capable, they are increasingly able to produce long and difficult
mathematical proofs, including solutions to long-standing open problems
\citep{openai2026tenadvances, alman2026truly}.
At the same time, it is becoming harder to verify
the correctness of these proofs. Existing verifiers, including outcome and process reward models,
were built for problems with a short checkable final answer \citep{lightman2023verify,
wang2023mathshepherd, rahman2025spark}, whereas a proof has no final answer to check; its
correctness lies in whether every step follows from the ones before it. Formal verification in a
proof assistant such as Lean offers an alternative, but translating an informal proof into a formal
language is itself difficult and requires expert-level effort. As a result, evaluating the proofs
that AI systems produce is hard, especially for non-experts, and reliable supervision for
training future systems is scarce.

A verifier trained on natural-language proofs could work well, but training it requires a
correctness label on every step, and such labels are scarce for open-ended proofs. 
Human annotation
is expensive even for short math problem solutions, as PRM800K needed about 800{,}000 step labels
\citep{lightman2023verify}. 
To address this challenge, we introduce LeanPRM, a framework that uses the Lean proof assistant \citep{demoura2021lean4}, an exact mechanical checker, to create correctness labels for informal math proofs. 
In LeanPRM, each natural-language
step is paired with a formal step; we use Lean to check the formal step, and the result becomes the label of
the corresponding natural-language step. 
Lean is used only to build the training data, and the trained verifier works directly with natural language.

\begin{figure}[t]
\centering
\includegraphics[width=\textwidth]{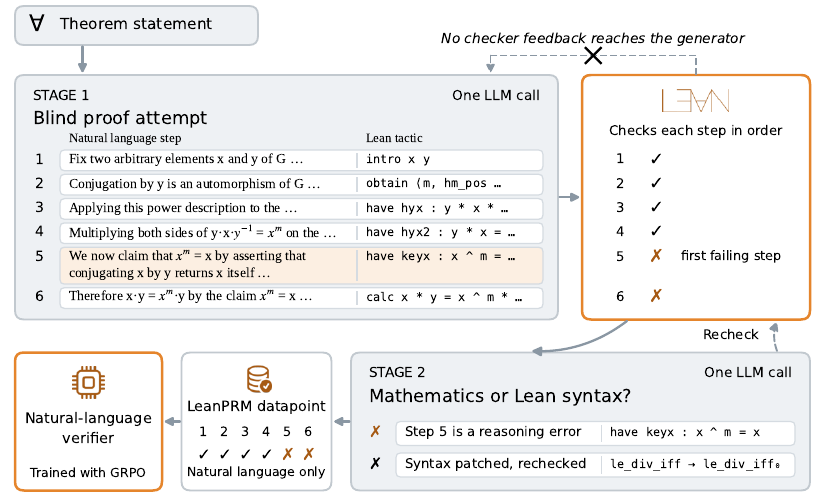}
\caption{Overview of LeanPRM, with one example from the corpus running through it. Top row: In Stage 1 an LLM writes a proof attempt from the theorem statement alone, where each natural-language step is paired with its formal step; we use Lean to check the step correctness in order. Bottom row: Stage 2 decides whether the first failure lies in the mathematics or only in the formal syntax (syntax failures are patched and rechecked once); the verifier model is trained on the natural-language steps and their labels only. Here step 5 asserts without justification that conjugation by $y$ fixes $x$, so steps 1 to 4 are labeled correct and steps 5 and 6 incorrect.}
\label{fig:pipeline}\label{fig:example}
\end{figure}

We explain how we build the data in Figure~\ref{fig:pipeline} and give the full procedure in
Section~\ref{sec:pipeline}. For each theorem statement we ask a model to prove it in a single pass,
writing each step in natural language together with its corresponding Lean step. The model never
sees the output of running the Lean step it generated. If it could, it would repair its mistakes or
give up, and we would lose the genuine errors we wish to label. As a result, the errors in LeanPRM
are the model's own, and we don't have to prompt it in order to induce them. Following this we run
Lean verification on the formal proof. Steps before the first failing Lean step are labeled correct, and the
failing step and every later step are labeled incorrect. A step can also fail for purely formal
reasons while the mathematics is sound, so we use a second model call to read the Lean error
message, repair syntax errors once and classify the failure. We keep only proofs whose first failure
is a mathematical error. In total we obtain 10{,}634 labeled proofs with 152{,}005 step labels and
no human annotation, which we release together with LeanPRM-Bench, a held-out benchmark of 200
proofs.

In our experiments, we use LeanPRM to train verifier models for natural language.  The results show
improved performance relative to the same training process applied to the human labels of PRM800K.
With the same model, training recipe and number of updates, LeanPRM-Verifier-4B beats the
PRM800K-trained verifier on four of five benchmarks, by  \newrev{14.4 }points of exact step-label accuracy on
LeanPRM-Bench and by  \newrev{9.7 and 9.4 }points on Hard2Verify and IMO-GradingBench. It also picks correct
proofs more often than random selection in best-of-$N$ on IMO-level problems, where the PRM800K
verifier stays close to chance. Training on LeanPRM improves verifiers at every size we tried, from
0.6B to 8B parameters, on LeanPRM-Bench, Hard2Verify and IMO-GradingBench. We provide detailed
results in Section~\ref{sec:experiments}.

We make three contributions.

\begin{itemize}
\item LeanPRM, to our knowledge the first framework that uses Lean to produce step-level correctness
labels for natural-language proofs at scale, and its corpus of 10{,}634 labeled proofs
(Section~\ref{sec:pipeline}).

\item LeanPRM-Bench, a held-out benchmark of 200 natural-language proofs with step labels, which we
release for evaluating proof verifiers (Section~\ref{sec:bench}).

\item A controlled evaluation showing that, with the same recipe and training budget, verifiers
trained on LeanPRM outperform verifiers trained on existing human-labeled verification data
on four of five
benchmarks, select better proofs at test time on IMO-level problems, and improve at every model size
from 0.6B to 8B on three of the five benchmarks (Section~\ref{sec:experiments}).
\end{itemize}

Because LeanPRM needs no human in the loop, the difficulty of its data is set by the prover rather than by the availability of expert annotators: as stronger provers write longer and harder proofs, the same pipeline can supervise stronger verifiers. 
More broadly, LeanPRM shows that a formal checker can supervise a model that never sees formal language at test time, so that synthetic data grounded in Lean transfers to informal reasoning. 
In future work, we hope to extend this formal-to-informal generalization beyond mathematics, to domains where parts of an argument can be checked mechanically even when the argument as a whole is written in natural language.

\section{LeanPRM: Lean-Grounded Process Supervision}\label{sec:pipeline}


We introduce LeanPRM, a framework for generating natural-language proofs with correctness labels on
every step, grounded in Lean. \revision{Lean is a proof assistant that checks formal proofs and tactics are commands used to
construct them \citep{demoura2021lean4}.}
The framework has two stages (Figure~\ref{fig:pipeline}): (i) a model
writes a natural-language proof together with a formal proof in Lean, pairing each natural-language
step with the corresponding Lean tactic, and (ii) Lean compiles the formal proof and a second model
reviews the result. If compilation failed, the second model decides whether the failure is a
mathematical error or only a syntax error, which it attempts to repair once. If compilation
succeeded, it checks that the natural-language steps state the same mathematics as the formal proof.
\revision{Up to the first failure, each natural-language step is labeled by the outcome of its
tactic. Every step subsequent to the first failure is labeled incorrect.}

\begin{figure}[t]
\centering
\includegraphics[width=\textwidth]{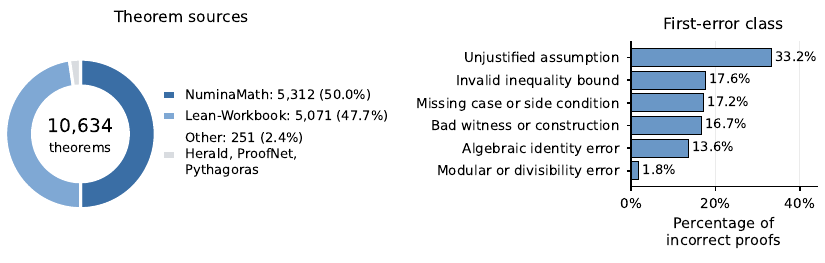}
\caption{Composition of LeanPRM. Left: sources of the 10{,}634 theorems. Right: error class of the
first failing step in each of the 4{,}264 incorrect proofs. \revision{Bars show percentages of incorrect proofs.} Every incorrect proof fails first at a
mathematical error, most often an unjustified assumption.}
\label{fig:data}
\end{figure}

\textbf{Stage 1: Proof generation.}\quad We start from about 17{,}800 distinct theorem statements
from public formal corpora, mostly from NuminaMath-LEAN \citep{wang2025kimina} and Lean-Workbook
\citep{ying2024leanworkbook,lin2025goedel} (Figure~\ref{fig:data}, with all sources listed in
Appendix~\ref{app:splits}). \revision{The seed statements span algebra, inequalities, number theory, and existence problems,
with some calculus and combinatorics. NuminaMath statements are selected from the ``hard'' tier
assigned by Gemini Flash-Lite, and Lean-Workbook is sampled by statement category without difficulty labels.}
For each statement, one model call writes a proof attempt as numbered
steps, each a sentence of ordinary mathematical prose paired with the formal step that carries it
out (prompt in Appendix~\ref{app:prompts}). The model sees only the statement and generates a proof
in one shot, without receiving feedback from the checker. The model is always asked to
prove the theorem and is never instructed to make a mistake, so every error in LeanPRM is one the
model made while genuinely trying to produce a proof. Because many statements were selected for
difficulty (Appendix~\ref{app:splits}), many of these genuine attempts fail. 
We used Claude Sonnet~5 for most theorems and Claude Opus~4.8 for an earlier subset.

\textbf{Stage 2: Verification and error classification.}\quad The frozen formal proof is compiled
with Lean~4 and Mathlib (both v4.25.0-rc2), and no compiler output is shown to the generator.
Because each sentence was written with its own formal step, a compilation error points to exactly
one step. A formal step can fail to compile even when the mathematics is right, so a second and
final model call reads the statement, the steps and the compiler output and returns a verdict on the
step. The verdict may be correct (the proof compiled and its natural language description is
faithful to it), mathematical error (the call names the first failing step and assigns it an error
class), or syntax error (the call gets a single attempt to patch the corresponding Lean step without
changing the mathematics, and the patched proof is kept only if it then compiles). At this stage,
the proof may also be discarded due to an ill-posed statement, malformed output, or prose and formal
proof that disagree. A proof that compiles is labeled correct at every step. Otherwise, the first
failing step and every step after it are labeled incorrect.
On steps that are true but inadequately justified, this
approach can differ from ProcessBench \citep{zheng2024processbench}, which counts a missing
justification as an error only when it affects the validity of the solution
(Appendix~\ref{sec:analysis}).

\subsection{LeanPRM Corpus}
\label{sec:data}

Stage~2 produces a labeled proof for 13{,}940 of these statements, and 13{,}007 remain after the
lexical filters and the removal of duplicate and empty proofs. After the quality control described
below, LeanPRM contains 10{,}634 natural-language proofs, 6{,}370 correct and 4{,}264 incorrect,
with 152{,}005 labeled steps, obtained with no human annotation. Of these, 98\% come from
NuminaMath-LEAN and Lean-Workbook. Table~\ref{tab:stats} in Appendix~\ref{app:splits} breaks the
corpus down by source and split.

\textbf{Error taxonomy.}\quad
\label{sec:taxonomy}
Stage 2 assigns each reasoning error one of ten classes. Six describe mathematical faults:
unjustified assumptions, invalid inequality bounds, wrong witnesses or constructions, missing cases
or side conditions, faulty algebraic identities, and errors in modular or divisibility reasoning.
The other four describe failures of the formalization rather than of the mathematics. Examples of
these errors include a lemma applied where its formal hypotheses do not hold, a rewrite in the wrong
direction, automation asked to close a goal it cannot, and a residual class for anything else.
Appendix~\ref{app:classes} gives the definitions the classifier works from.

\begin{wraptable}{r}{0.46\textwidth}
\centering
\footnotesize
\caption{\revision{Blind audit of flagged steps, 30 per class (all formalization-side
classes and the two largest mathematical ones). \emph{Judged sound}: correct
and justified. \emph{Share of errors}: fraction of incorrect proofs in the
unfiltered pool. Formalization-side steps are mostly sound. Mathematical
steps are mostly not sound.}}
\label{tab:audit}
\setlength{\tabcolsep}{4pt}
\begin{tabular}{lcc}
\toprule
First-error class & \makecell{Judged\\sound} & \revision{\makecell{Share of\\errors}} \\
\midrule
\multicolumn{3}{l}{\emph{Formalization-side}} \\
\quad Wrong rewrite direction  & $28/30$ & $2.7\%$ \\
\quad Theorem misapplication   & $26/30$ & $9.2\%$ \\
\quad Other                    & $26/30$ & $9.5\%$ \\
\quad Automation misuse        & $23/30$ & $13.4\%$ \\
\midrule
\multicolumn{3}{l}{\emph{Mathematical}} \\
\quad Unjustified assumption   & $13/30$ & $21.3\%$ \\
\quad Invalid inequality bound & $13/30$ & $11.3\%$ \\
\bottomrule
\end{tabular}
\vspace{-8pt}
\end{wraptable}

\textbf{Quality control.}\quad
\label{sec:audit}
\revision{Because the corpus has no human annotations, we checked the automatic Stage~2 labels with two blind
audits using Claude Opus~5 as the judge.}
The judge never saw our labels (prompts in Appendix~\ref{app:prompts}). In the first audit, the judge read 100
stratified proofs from the unfiltered pool and named the first incorrect step, if any. It returned a
usable answer for 99 of them. It disputed 1 of the 52 proofs we label correct. Of the 47 we label
incorrect, it found an error in 27 and none in 20. The second audit shows where that disagreement
comes from (Table~\ref{tab:audit}). Among 30 flagged steps per class, steps in the
formalization-side classes, 35.8\% of the incorrect attempts in the pool, were judged sound in 77\%
to 93\% of cases, while steps in the mathematical classes were judged sound less than half the time.
We therefore remove every attempt whose first failure is formalization-side, so every negative in
LeanPRM carries a mathematical error class. We additionally use lexical filters to make sure the
natural language proofs do not reference Lean and remove incorrect proofs whose failing step states
its own error (Appendix~\ref{app:filter}). Mathematical steps are judged sound in 13 of 30 cases because these are mostly correct steps whose Lean step failed for formal reasons and that Stage~2 classified as mathematical errors, as noted in Section~\ref{sec:conclusion}.

\subsection{LeanPRM-Bench}
\label{sec:bench}

LeanPRM-Bench is a held-out set of 200 theorems, 100 with a correct proof and 100 with an incorrect
one. \revision{No benchmark theorem appears in the training data. A proof is labeled correct if its Lean code compiles
and a reviewing model confirms that the natural-language steps match the formal proof.} Every incorrect proof first fails at a
mathematical error (Appendix~\ref{app:splits}). The benchmark is hard for current models: prompted
with the same instructions, four commercial models (Claude Sonnet~5, Claude Opus~5, Claude Fable~5.1 and GPT-5.6 Sol) reach \revision{71.7 to 80.3 mean exact step-label accuracy over three
independent runs (Table~\ref{tab:main}). In their original reference runs, all four mislabel the same 26 proofs (Appendix~\ref{app:benchmark}).} We report exact step-label accuracy, the fraction of proofs
labeled correctly at every step (Table~\ref{tab:main}).
We release LeanPRM-Bench with its step labels, error classes and formal proofs.

\section{Experiments}
\label{sec:experiments}


\textbf{Training.}\quad The verifier reads a theorem and its numbered steps, writes its reasoning,
and then gives each step a label, correct or incorrect, and the proof an overall verdict (prompt in
Appendix~\ref{app:prompts}). We train Qwen3 models \citep{yang2025qwen3} with 0.6B, 4B and 8B
parameters, in thinking mode, with group relative policy optimization \citep[GRPO,
][]{shao2024deepseekmath}. For each proof we sample eight responses at temperature 1, normalize
their rewards within the group, and use the normalized reward as the advantage of every token in the
response. The importance ratio is clipped to $[0.8, 1.26]$, and a KL penalty with coefficient
$10^{-3}$ keeps the policy close to the initial model. Each update uses 64 proofs and a learning
rate of $10^{-6}$. Every trained model in the main comparison gets  \newrev{500 }updates with this
configuration. The PRM800K baseline changes the training data. The NuminaMath solver also replaces
the verification reward with a final-answer reward. Those  \newrev{500 updates cover 32}{,} \newrev{000 proofs: 3.12
}passes over the 10{,}264 LeanPRM training proofs and  \newrev{0.45 }over the 70{,}502 PRM800K training proofs.
Appendix~\ref{app:hparams} lists every hyperparameter.

\textbf{Reward.}\quad Let $R_{\mathrm{c}}$ be the fraction of truly correct steps that the verifier
labels correct and $R_{\mathrm{i}}$ the fraction of truly incorrect steps that it labels incorrect,
each set to 1 when there are no such steps. With predicted verdict $\hat{v}$ and true verdict $v$,
the reward is
\begin{gather}
r = 0.7\,\cdot R_{\mathrm{step}} + 0.3\,\cdot \mathbf{1}[\hat{v} = v], \nonumber\\
R_{\mathrm{step}} =
\begin{cases}
R_{\mathrm{c}}\,R_{\mathrm{i}} & \text{if the proof contains an error,}\\
\mathbf{1}[\text{every step labeled correct}] & \text{otherwise.}
\end{cases}
\label{eq:reward}
\end{gather}
On a proof containing both correct and incorrect steps, the product rules out both
constant-label shortcuts. Labeling every step correct gives $R_{\mathrm{i}}=0$, and labeling
every step incorrect gives $R_{\mathrm{c}}=0$. If the first step is incorrect, all steps are
incorrect under our labeling convention and the empty-class recall is 1. For the proof in
Figure~\ref{fig:pipeline}, whose steps 5 and 6 are incorrect, a verifier that flags only step 6 gets
$R_{\mathrm{c}}=1$ and $R_{\mathrm{i}}=1/2$, so $R_{\mathrm{step}}=0.5$ and, with the verdict
necessarily incorrect, $r=0.65$. The reward is 0 when the output is malformed, has no reasoning
section, or gives a verdict that contradicts its own step labels.

\textbf{Evaluation.}\quad We evaluate on LeanPRM-Bench and on four external benchmarks, each on its
full evaluation set, apart from the 13 context-length exclusions on IMO-GradingBench described in Appendix~\ref{app:eval}, and with one metric per benchmark. We use Hard2Verify \citep{pandit2025hard2verify} (response-level
balanced accuracy), ProcessBench \citep{zheng2024processbench} (F1, the harmonic mean of accuracy on
errors and correct solutions, averaged over its four subsets, 3{,}400 test cases), PRMBench
\citep{song2025prmbench} (first-error accuracy over 6{,}216 rows, not its default PRMScore, which marks only the erroneous steps as incorrect and so would penalize every model for following our instruction to label all steps after the first error incorrect) and
IMO-GradingBench \citep{luong-etal-2025-towards} (balanced accuracy on the grade binarized to fully
correct or not, instead of its four-way accuracy, since our verifiers give a binary verdict and only
about 30\% of its proofs are graded correct). All models get the same prompt layout and label parser, with hosted-response formatting
normalization described in Appendix~\ref{app:eval}. A response that fails to parse counts as wrong. We report means over three runs, using decoding seeds 42, 43, and 44 at temperature 1
for open-weight models and independent hosted runs for commercial models (Appendix~\ref{app:eval}).

\textbf{Baselines.}\quad We compare with the untrained Qwen3-4B and with two models trained by the
same recipe on other data: one on PRM800K \citep{lightman2023verify}, converted to our step-label
format (PRM800K-Verifier-4B), and one with reinforcement learning on solving 19{,}200 NuminaMath
problems with a final-answer reward (NuminaMath-Solver-4B), which tests whether verification ability
comes from mathematical reinforcement learning alone. For reference we also prompt four commercial
models with the same instructions: Claude Sonnet~5, Opus~5 and Fable~5.1 through the Anthropic API
with extended thinking, and GPT-5.6 Sol through the Codex command-line interface with every tool
disabled (Appendix~\ref{app:eval}). Both trained baselines change little on their own validation
data over the second half of their  \newrev{500 }updates (Figure~\ref{fig:curves}, middle), so the  \newrev{500 }step
budget is a reasonable setting.

\subsection{Results}
\label{sec:results}

\begin{table}[t]
\centering
\small
\setlength{\tabcolsep}{4pt}
\begin{tabular}{@{}lccccc@{}}
\toprule
& LeanPRM-Bench & Hard2Verify & ProcessBench & PRMBench & IMO-GB \\
& exact & response BA & avg.\ F1 & first err. & BA \\
\midrule
Qwen3-4B (initial policy) & 48.2\,{\scriptsize$\pm$1.3} & 37.1\,{\scriptsize$\pm$4.5} &
74.3\,{\scriptsize$\pm$0.5} & 42.5\,{\scriptsize$\pm$0.5} & 53.8\,{\scriptsize$\pm$0.1} \\
PRM800K-Verifier-4B & \newrev{60.3\,{\scriptsize$\pm$1.6} }& \newrev{63.1\,{\scriptsize$\pm$0.6} }& \textbf{\newrev{83.0}}\,{\scriptsize$\pm$0.3} & \newrev{43.3\,{\scriptsize$\pm$0.3} }& \newrev{57.0}\,{\scriptsize$\pm$0.4} \\
NuminaMath-Solver-4B & \newrev{48.3}\,{\scriptsize$\pm$1.0} & \newrev{33.6\,{\scriptsize$\pm$3.4} }& \newrev{71.2}\,{\scriptsize$\pm$0.9} & \newrev{41.5\,{\scriptsize$\pm$0.2} }& \newrev{52.4}\,{\scriptsize$\pm$0.4} \\
\rowcolor{ourrow}
LeanPRM-Verifier-4B (ours) & \textbf{\newrev{74.7}}\newrev{\,{\scriptsize$\pm$1.8} }& \textbf{\newrev{72.8}}\newrev{\,{\scriptsize$\pm$0.9} }& \newrev{74.9\,{\scriptsize$\pm$0.4} }& \textbf{\newrev{47.4}}\newrev{\,{\scriptsize$\pm$0.2} }& \textbf{\newrev{66.4}}\newrev{\,{\scriptsize$\pm$1.1} }\\
\midrule
\rowcolor{ourrow}
LeanPRM-Verifier-4B$^{\dagger}$ & 78.7\,{\scriptsize$\pm$1.3} & 79.0\,{\scriptsize$\pm$0.7} &
72.6\,{\scriptsize$\pm$0.3} & 48.3\,{\scriptsize$\pm$0.2} & 71.8\,{\scriptsize$\pm$1.0} \\
Claude Sonnet 5 & 71.7\,{\scriptsize$\pm$0.2} & 83.1\,{\scriptsize$\pm$0.4} & 84.9\,{\scriptsize$\pm$0.4} & 48.6\,{\scriptsize$\pm$0.1} & 80.7\,{\scriptsize$\pm$0.4} \\
Claude Opus 5 & 80.3\,{\scriptsize$\pm$1.2} & 81.0\,{\scriptsize$\pm$0.2} & 84.5\,{\scriptsize$\pm$0.1} & 47.6\,{\scriptsize$\pm$0.2} & 81.8\,{\scriptsize$\pm$0.3} \\
Claude Fable 5.1 & 78.3\,{\scriptsize$\pm$0.6} & 86.0\,{\scriptsize$\pm$0.7} & 84.7\,{\scriptsize$\pm$0.0} & 48.6\,{\scriptsize$\pm$0.1} & 85.4\,{\scriptsize$\pm$0.8} \\
GPT-5.6 Sol & 75.3\,{\scriptsize$\pm$1.2} & 85.5\,{\scriptsize$\pm$1.4} & 79.8\,{\scriptsize$\pm$0.2} & 47.1\,{\scriptsize$\pm$0.3} & 83.2\,{\scriptsize$\pm$0.4} \\
\bottomrule
\end{tabular}
\vspace{0.3cm}
\caption{Verification accuracy. Open-weight models use temperature 1. Commercial-model settings
are given in Appendix~\ref{app:eval}. LeanPRM-Bench is scored by exact
step-label accuracy (Section~\ref{sec:bench}), and the other columns use the metrics of
Section~\ref{sec:experiments}. The top block compares 4B models trained for \newrev{500 }updates with the
same recipe, with the untrained model for reference, and bold marks the best value in that block.
The bottom block holds reference rows: $^{\dagger}$a separate 4B run of 1{,}582 updates, only the
last 300 of them on LeanPRM (Appendix~\ref{app:training}), and commercial models prompted
rather than trained for the task. At equal training, LeanPRM-Verifier-4B is best on four of the five
benchmarks. Entries report mean $\pm$ population standard deviation over three evaluation runs
(Appendix~\ref{app:eval}).}
\label{tab:main}
\end{table}

\begin{figure}[t]
\centering
\includegraphics[width=\textwidth]{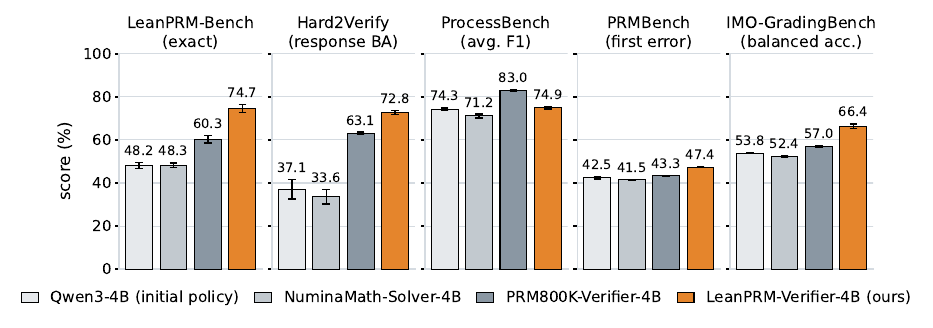}
\caption{The top block of Table~\ref{tab:main} per benchmark: 4B models trained for \newrev{500 }updates with
the same recipe, and the untrained model. Bars are means over three decoding seeds with one standard
deviation as error bars. LeanPRM-Verifier-4B is best on four
of the five benchmarks and PRM800K-Verifier-4B on ProcessBench.}
\label{fig:main}
\end{figure}

\textbf{Process verification benchmarks.} \quad Table~\ref{tab:main} and Figure~\ref{fig:main} give
the main comparison. At the same budget, LeanPRM-Verifier-4B beats PRM800K-Verifier-4B on four of
the five benchmarks: by  \newrev{14.4 }points on LeanPRM-Bench,  \newrev{9.7 }on Hard2Verify,  \newrev{4.1 }on PRMBench and  \newrev{9.4 }on
IMO-GradingBench. PRM800K-Verifier-4B keeps a  \newrev{8.1}-point lead on ProcessBench, whose short solutions
come from the same distribution as its training data and whose labels, unlike ours, count a missing
justification as an error only when it affects the validity of the solution
(Appendix~\ref{sec:analysis}). Reinforcement learning on solving does not produce a verifier:
NuminaMath-Solver-4B raises final-answer accuracy on its own validation problems from 30.0\% to
 \newrev{47.5}\%, yet stays within a point of the untrained model on LeanPRM-Bench and falls below it on every
external benchmark. We additionally report results for a longer LeanPRM training run denoted as
LeanPRM-Verifier-4B$^{\dagger}$: 532 updates on proofs from an earlier joint-generation procedure,
750 on the unfiltered pool and 300 on LeanPRM (Appendix~\ref{app:training}). Its data differ as well
as its length, so we report it only for reference. We report results for the commercial models as a
reference too. They lead on Hard2Verify, ProcessBench and IMO-GradingBench, and
Appendix~\ref{app:benchmark} discusses the 26 LeanPRM-Bench proofs that all four mislabel
in their original reference runs. Standard
deviations across seeds are at most 3.2 points for the trained verifiers across decoding seeds.

\begin{figure}[t]
\centering
\includegraphics[width=\textwidth]{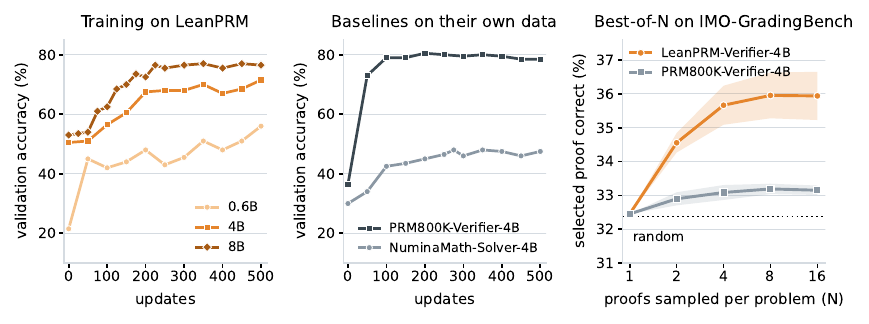}
\caption{Training curves and test-time selection. Left: exact step-label accuracy on the LeanPRM validation split during training, for the verifiers trained from Qwen3 base models of three sizes; the 4B point at \newrev{update }300 was measured \newrev{separately }with the same greedy protocol. Middle: the two trained baselines on their own validation splits, PRM800K-Verifier-4B by exact accuracy and NuminaMath-Solver-4B by final-answer accuracy. Right: best-of-$N$ selection on IMO-GradingBench, the share of selected proofs graded correct, averaged over the 30 problems and three decoding seeds (shaded band: one standard deviation); the dotted line is random selection. LeanPRM-Verifier-4B picks better proofs as $N$ grows, while PRM800K-Verifier-4B stays near chance.}
\label{fig:bestofn}\label{fig:curves}
\end{figure}

\textbf{Test-time scaling.}\quad 
\label{sec:bestofn}
A verifier is useful at test time if it can pick a good proof out of several attempts. We measure
this on IMO-GradingBench, which has about 33 human-graded proofs for each of its 30 problems. For
each problem we draw $N$ proofs, let the verifier pick one it accepts (at random among those it
accepts, or among all of them if it accepts none), and record whether that proof is graded correct.
Figure~\ref{fig:bestofn} (right) shows the result. At $N=16$, LeanPRM-Verifier-4B picks a correct
proof  \newrev{35.9}\% of the time, against 32.4\% for random selection, and its advantage grows with $N$ \newrev{up to $N=8$}.
PRM800K-Verifier-4B, trained with the same recipe and budget, reaches  \newrev{33.2}\%: it accepts about  \newrev{79}\%
of all proofs although only 30\% are graded correct, so its picks differ little from chance.

\begin{figure}[t]
\centering
\includegraphics[width=\textwidth]{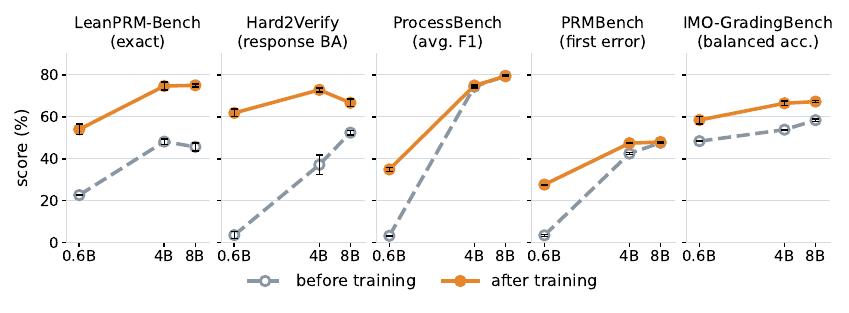}
\caption{Accuracy before and after training on LeanPRM, by model size. Each panel is one benchmark;
open circles are the untrained Qwen3 models and filled circles the trained verifiers after \newrev{500
}updates. Training improves every model on LeanPRM-Bench, Hard2Verify and IMO-GradingBench; the
untrained 0.6B model scores near zero on three benchmarks because it rarely produces well-formed
labels. Points and error bars show the mean and population standard deviation over three decoding seeds.}
\label{fig:scaling}
\end{figure}

\textbf{Model scale.}\quad 
\label{sec:scale}
We compare verifiers trained from Qwen3 base models of 0.6B, 4B and 8B parameters in
Figure~\ref{fig:scaling}, with the numbers in Appendix~\ref{app:scale}. Training helps at every size
on LeanPRM-Bench, Hard2Verify and IMO-GradingBench. After  \newrev{500 }updates the gain on LeanPRM-Bench over
the untrained model is  \newrev{31 }points at 0.6B,  \newrev{27 }at 4B and  \newrev{29 }at 8B. The untrained 8B model starts 2.5
points below the untrained 4B model on LeanPRM-Bench (45.7 against 48.2), so its larger gain only
brings it level with the 4B model. The 0.6B model nevertheless stays far behind. Before training it
produces well-formed labels for fewer than one proof in five on Hard2Verify, ProcessBench and
PRMBench, and after training it is  \newrev{21 }points below the 4B model on LeanPRM-Bench. The 4B and 8B
models reach \newrev{nearly }the same LeanPRM-Bench accuracy,  \newrev{74.7 and 75.0}, and are within  \newrev{0.8 }points of each other on
 PRMBench and IMO-GradingBench; the \newrev{8B model is 4.4 points ahead on ProcessBench and the }4B model is ahead on Hard2Verify. Formatting failures contribute to that gap. The 8B model returns an unparsable answer for  \newrev{9}\% of Hard2Verify proofs,
against  \newrev{2}\% for the 4B model, and on the answers that parse the two score  \newrev{74.7 and 75.1}. On
ProcessBench and PRMBench, the 8B model starts highest (79.5 and 47.6) and ends  \newrev{within 0.4 points of
}its untrained scores . Figure~\ref{fig:curves} (left) shows validation accuracy during training for
the three sizes.

\section{Related Work}

\begin{table}[t]
\centering
\small
\footnotesize
\renewcommand{\arraystretch}{0.92}
\setlength{\tabcolsep}{4pt}
\begin{tabular}{@{}llcccc@{}}
\toprule
Method & Step labels from & \makecell{No human\\step labels} & \makecell{Open-ended\\proofs} &
\makecell{Error\\classes} & \makecell{Trained\\verifier} \\
\midrule
PRM800K \citep{lightman2023verify} & Human annotators & \xmark & \xmark & \xmark & \cmark \\
Math-Shepherd \citep{wang2023mathshepherd} & Rollouts & \cmark & \xmark & \xmark & \cmark \\
OmegaPRM \citep{luo2024improve} & Rollouts (tree search) & \cmark & \xmark & \xmark & \cmark \\
FoVer \citep{kamoi2025fover} & Z3, Isabelle & \cmark & \xmark & \xmark & \cmark \\
Safe \citep{liu2025safe} & Lean~4 & \cmark & \xmark & \xmark & \xmark \\
TP-as-a-Judge \citep{leang2025tpjudge} & Lean~4 (per solution) & \cmark & \xmark & \xmark & \xmark \\
LeanTutor \citep{patel2026leantutor} & Lean~4 & \cmark & \cmark & \cmark & \xmark \\
\midrule
\rowcolor{ourrow}
\textbf{LeanPRM (ours)} & Lean~4 & \cmark & \cmark & \cmark & \cmark \\
\bottomrule
\end{tabular}
\vspace{0.3cm}
\caption{Sources of step-level labels in prior work and in LeanPRM. For each method we give the
origin of its step labels and whether they avoid human annotation, cover open-ended
natural-language proofs, give each failing step a mathematical error type, and train a verifier
that reads the proof text. Only LeanPRM does all four.}
\label{tab:comparison}
\end{table}

\revision{LeanPRM draws on two lines of work.} Process supervision trains
verifiers on step-level labels, which mostly come from human annotators or from the outcomes of
sampled continuations. Formal mathematics offers exact checking, which has mainly been used to train
and evaluate provers that write Lean. We connect the two by using Lean only to label
natural-language proofs, and the verifier we train never sees a formal proof.

\textbf{Process supervision.}\quad \citet{lightman2023verify} scaled up process supervision for
mathematical reasoning with human step-level annotations. Subsequent work reduces this annotation
cost by deriving process supervision from outcome signals: Math-Shepherd estimates step quality from
the correctness of sampled continuations \citep{wang2023mathshepherd}, OmegaPRM scales such
annotation using tree search \citep{luo2024improve}, and RL Tango trains a generative process
verifier directly from outcome-level rewards \citep{zha2025rltango}. Outcome-trained verifiers were
introduced for word problems by \citet{cobbe2021verifiers}, and \citet{uesato2022process} ran the
first comprehensive comparison of process- and outcome-based feedback on a natural-language task;
later work scales automated step labels through rollouts and value estimates
\citep{lu2024autopsv,setlur2024pav}, casts verification as generation \citep{zhang2024genrm}, and
distills practical lessons for training process reward models \citep{zhang2025prmlessons}, and such
verifiers are what makes test-time search effective \citep{snell2024testtime,ma2023stepreward},
since models struggle to correct their own reasoning without external feedback
\citep{huang2023selfcorrect}.

These approaches obtain labels from the model-generated solution distribution, either through human
judgment or through rollout statistics. We instead obtain them from an external decision procedure,
which makes the correctness criterion explicit and independent of the model being supervised. In
this direction, \revision{FoVer \citep{kamoi2025fover} is closely related. It trains process reward
models on formal reasoning traces labeled with Z3 and Isabelle, then evaluates their transfer to
natural-language reasoning. LeanPRM instead trains directly on natural-language competition proofs
paired with Lean tactics and distinguishes mathematical errors from formalization failures.}
Safe \citep{liu2025safe} checks each step of a
natural-language solution in Lean~4 to score solutions at inference, and Theorem Prover as a Judge
\citep{leang2025tpjudge} uses Lean verdicts on whole solutions to select fine-tuning data.

\textbf{Formal mathematics.}\quad Lean~4 \citep{demoura2021lean4} and Mathlib \citep{mathlib2020}
have been widely used to construct data and train formal theorem provers
\citep{yang2023leandojo,lin2025goedel} and autoformalizers \citep{gao2024herald}. Provers such as
DeepSeek-Prover \citep{xin2024deepseekprover15,ren2025deepseekprover2}, Kimina-Prover
\citep{wang2025kimina}, Goedel-Prover-V2 \citep{lin2025goedelv2} and Seed-Prover
\citep{chen2025seedprover} learn from proof-assistant feedback, Lean-STaR interleaves informal
thoughts with tactics \citep{lin2024leanstar}, and miniF2F and PutnamBench track their progress
\citep{zheng2021minif2f,tsoukalas2024putnambench}. Autoformalization \citep{wu2022autoformalization}
and formalized problem sets such as Lean Workbook \citep{ying2024leanworkbook} supply the
statements, informal proofs can be mapped to formal proof sketches \citep{jiang2022dsp}, and
\citet{yang2024formalfrontier} survey the field.

Recent work also uses formal verification to evaluate natural-language reasoning
\citep{zhou2025step}, to check key steps during inference \citep{ospanov2025hermes}, or to provide
training signals for natural-language reasoning \citep{cao2026interleaved} and formal proof
generation \citep{kim2026process}. DeepTheorem uses automatically verifiable theorem outcomes
\citep{zhang2025deeptheorem}, while DeepSeekMath-V2 trains natural-language provers with a learned
proof verifier \citep{shao2025deepseekmath}. LeanTutor \citep{patel2026leantutor} checks student
proofs step by step in Lean to give feedback. \citet{ananthula2026decodable} train language models so that their written explanations carry a verifier's verdict, such as whether Lean accepts a proof, and find that the explanations can carry this information and still not reflect how the model reached its answer. Our work instead uses such step labels, at scale, as
training data for a verifier that reads only natural language, providing supervision that is more
scalable than expert annotation and more mechanically grounded than model judgments. 
Our use of the checker differs in that the object being supervised is natural-language reasoning:
the formal proof is never the model's output and never appears in its input.

\textbf{Verifier training and evaluation.}\quad Our verifiers are trained with group-relative policy
optimization \citep{shao2024deepseekmath}, a variant of PPO \citep{schulman2017ppo} that DeepSeek-R1
scaled to reasoning \citep{deepseek2025r1}, implemented in VERL \citep{sheng2024hybridflow}. The
prover-verifier framing follows \citet{anil2021pvg}, as applied to language models by
\citet{kirchner2024pvg}, and critic models that surface errors in model outputs
\citep{mcaleese2024critics} are the closest analog outside mathematics.

Evaluation of step-level verifiers has been addressed by ProcessBench \citep{zheng2024processbench},
which measures identification of the first erroneous step in short solutions to grade-school,
competition and Olympiad-level problems, including GSM8K and MATH
\citep{cobbe2021verifiers,hendrycks2021math}, PRMBench \citep{song2025prmbench}, which probes
specific failure modes of process reward models, and Hard2Verify \citep{pandit2025hard2verify},
which targets proofs whose correctness is difficult to establish. Appendix~\ref{sec:analysis}
discusses how a benchmark's annotation convention can interact with the convention used in training.

\section{Conclusion}
\label{sec:conclusion}

We introduced LeanPRM, a framework that uses Lean to label every step of natural-language proofs,
trained verifiers on its corpus of 10{,}634 proofs, and released a held-out benchmark,
LeanPRM-Bench. With the same recipe and \revision{training budget}, these verifiers outperform verifiers
trained on the human labels of PRM800K on four of five benchmarks, and they pick better proofs at
test time. A formal tool can therefore be used to train verifiers for informal reasoning, without
autoformalization. Lean is needed only once, offline, to build the training data, and the verifier
it produces checks natural language directly.

\textbf{Limitations.}\quad LeanPRM separates failures of the formal encoding from errors in the
mathematics, and the audit in Section~\ref{sec:audit} shows that the classes it removes consist
mostly of sound steps. The separation itself relies on a model's classification, checked with a
single judge model on a sample, so some borderline steps may still carry the wrong label. We
compared Lean labels with the human labels of PRM800K but did not train on the same proofs labeled
by an LLM judge, so we cannot yet say how much of the gain comes from the labels and how much from
training on proof-style data. LeanPRM-Bench comes from the same pipeline and the same Claude
generators as the training data, so the four external benchmarks carry most of the evidence that the
verifiers generalize. \revision{Because negative examples come from Claude's own mistakes, they may cover a narrower
range of errors than a proof verifier should detect.}

\textbf{Future work.}\quad We believe the same methodology for generating verifier training data
could carry over to other open-ended tasks that have no final answer but contain parts an exact
checker can verify, such as test suites for code or simulators for science.
Within mathematics, generating new theorem statements would let the corpus grow beyond existing
formal collections in size and difficulty. Verifiers trained on LeanPRM could also serve as
reward models, for training provers and for giving step-level feedback on proofs too hard for people
to check, which becomes more important as models approach and exceed human expertise.

\section{Acknowledgments}

We thank Tengyu Ma, Yuhong Geng, Siyan Zhao, and Liwei Jiang for helpful discussions, feedback, and suggestions throughout the development of this project. This work was funded in part by the DARPA expMath program (Agreement No. HR0011262E029). Pavel Izmailov and Salman Rahman were supported by a grant from the Alignment Project, funded by the UK AI Security Institute (grant APS2-100141). Saadia Gabriel was supported by a Google ML \& Systems Junior Faculty Award. Any opinions, findings, conclusions, or recommendations expressed in this material are those of the authors and do not necessarily reflect the views of DARPA, or the authors' affiliated institutions.

\bibliographystyle{plainnat}
\bibliography{iclr2027_conference}
\appendix
\section*{Appendix}

Appendix~\ref{app:classes} defines the ten error classes of Section~\ref{sec:taxonomy}, and
Appendix~\ref{app:splits} gives the data sources, the lexical filters and the splits behind
Section~\ref{sec:data}. Appendix~\ref{app:benchmark} examines the LeanPRM-Bench proofs that the
commercial models mislabel (Section~\ref{sec:results}). Appendices~\ref{app:training}
and~\ref{app:eval} give the training and evaluation details of Section~\ref{sec:experiments}, and
Appendix~\ref{sec:analysis} discusses the ProcessBench result. Appendix~\ref{app:scale} gives the
model-scale numbers behind Figure~\ref{fig:scaling}. Appendix~\ref{app:examples} shows further corpus examples,
Appendix~\ref{app:prompts} lists every prompt, and Appendix~\ref{app:hparams} gives the
hyperparameters.

\section{Error classes}
\label{app:classes}

Stage 2 assigns each reasoning error one of ten classes, defined in its instructions as follows. The
mathematical classes are: algebraic identity error, a false or invalid algebraic manipulation;
invalid inequality bound, a bound with the wrong direction or without justification; bad witness or
construction, a wrong witness for an existential claim or an invalid construction; modular or
divisibility error, incorrect modular arithmetic or divisibility reasoning; missing case or side
condition, a required case or side condition that is omitted; and unjustified assumption, a claim
asserted without support from the preceding steps. The formalization-side classes are: theorem
misapplication, a lemma applied where its hypotheses do not hold as stated; wrong rewrite direction,
an invalid or reversed rewrite or substitution; automation misuse, a claim that automation
discharges a goal it cannot; and other, a failure not captured by the classes above. Attempts whose
failure was recorded without a class are reported as unclassified.

\section{Data sources, filtering and splits}
\label{app:splits}

Theorem statements come from a difficulty-filtered subset of NuminaMath-LEAN \citep{wang2025kimina},
Lean-Workbook \citep{ying2024leanworkbook,lin2025goedel}, Pythagoras \citep{leang2026pythagoras},
Herald \citep{gao2024herald}, ProofNet \citep{azerbayev2023proofnet} and FormalQualBench
\citep{mathinc2026formalqualbench}. Table~\ref{tab:stats} gives the composition of LeanPRM by source
and by split.

\begin{table}[h]
\centering
\caption{Size of LeanPRM by source (top) and by split (bottom); each theorem contributes one proof. NuminaMath supplies most of the incorrect proofs and Lean-Workbook most of the correct ones.}
\label{tab:stats}
\begin{tabular}{@{}lrrr@{}}
\toprule
& Correct & Incorrect & Total \\
\midrule
NuminaMath & 1{,}334 & 3{,}978 & 5{,}312 \\
Lean-Workbook & 4{,}787 & 284 & 5{,}071 \\
Pythagoras & 230 & 1 & 231 \\
Herald & 13 & 1 & 14 \\
ProofNet & 6 & 0 & 6 \\
\midrule
Training & 6{,}170 & 4{,}094 & 10{,}264 \\
Validation & 100 & 70 & 170 \\
LeanPRM-Bench & 100 & 100 & 200 \\
\midrule
LeanPRM & 6{,}370 & 4{,}264 & 10{,}634 \\
\bottomrule
\end{tabular}
\end{table}

Theorem identity is defined modulo the declaration name, and every split is drawn by theorem group
so that alpha-renamed restatements cannot cross a boundary. The training split holds 10{,}264
theorems, 6{,}170 with a correct proof and 4{,}094 with an incorrect one; within each source the
smaller verdict cell is oversampled to parity, giving a training view of 17{,}670 rows. The
validation split holds 170 theorems (100 correct, 70 incorrect) and LeanPRM-Bench 200, 100 of each
verdict. LeanPRM-Bench was fixed before the class-based filter of Section~\ref{sec:audit}, and 30 of
its incorrect proofs turned out to have a formalization-side first error. We replaced them with 30
incorrect proofs from the validation split whose first error is mathematical, which is why the
validation split holds 170 theorems, and re-evaluated every system on the replaced items. The
validation curves in Figure~\ref{fig:curves} were recorded during training on the original
200-theorem validation split, which still contained those 30 proofs. No model was trained on them,
and none of the models compared at  \newrev{500 }updates was selected on validation.
LeanPRM-Verifier-4B$^{\dagger}$ is the exception. Its second stage was stopped at the best
checkpoint on an earlier validation split that held 17 of these 30 proofs.

\paragraph{Lexical filters.}\label{app:filter} Two lexical checks guard the prose. During
generation, an attempt is regenerated if any natural-language step uses one of the following terms
as a whole word, matched case-insensitively: lean, tactic, compiler, syntax, solver, automation,
automated, verifier, engine, linarith, nlinarith, field\_simp, norm\_num, aesop, omega. No proof in
LeanPRM contains any of them. When the splits are built, an incorrect proof is removed if its first
failing step announces its own error with phrases such as unjustified, invalid, false or does not
follow, since such text would give the label away. This second check removed 55.2\% of the incorrect
proofs from the earlier joint procedure and 2.3\% of those from the blind procedure. The Lean lines
written in Stage~1 are not filtered for sorry, admit or native\_decide, which appear in 605 proofs
of LeanPRM. Lean accepts \texttt{sorry}/\texttt{admit} as placeholders. \texttt{native\_decide} is not one.
A lexical audit found one positive training proof with explicit \texttt{sorry}, retained in the
experiments, and no explicit placeholders in positive validation or benchmark proofs.
Transitive dependencies were not checked. Only the repairs made in Stage~2 are barred from using any of the three.

\section{Proofs the frontier models misjudge}
\label{app:benchmark}

\begin{figure}[t]
\centering
\includegraphics[width=\textwidth]{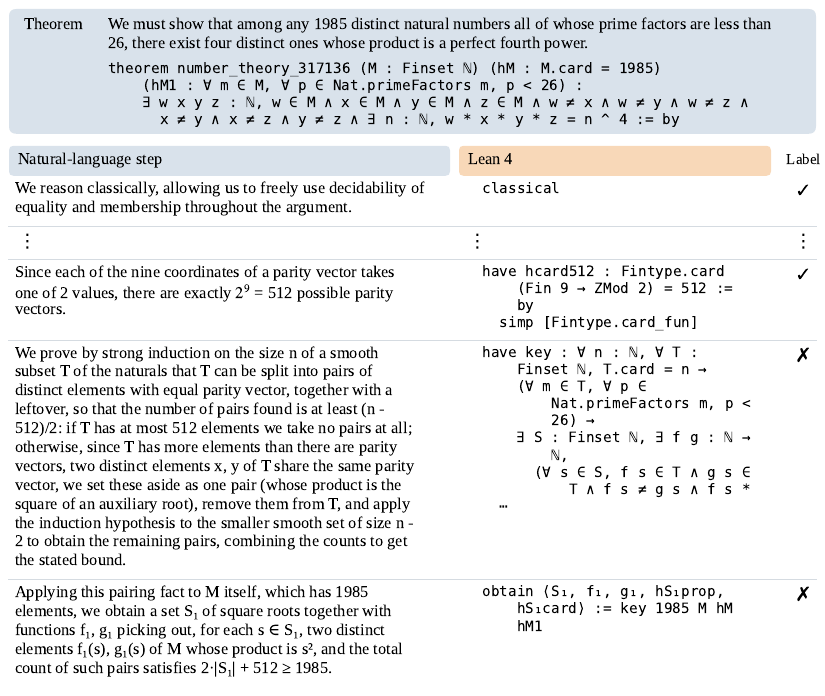}
\caption{A proof the trained verifier labels exactly and no commercial system does. Step 5 asserts,
by an unelaborated strong induction, that any smooth set of $n$ numbers yields at least $(n-512)/2$
disjoint pairs with equal parity vectors; the argument as written does not justify the count, the
checker fails at this step, and the class is unjustified assumption. Sonnet~5, Fable~5.1 and
PRM800K-Verifier-4B accept every step, Opus~5 and GPT-5.6 Sol first object at step 6, and
LeanPRM-Verifier-4B$^{\dagger}$ flags step 5.}
\label{fig:showcase_pairing}
\end{figure}

\begin{figure}[t]
\centering
\includegraphics[width=\textwidth]{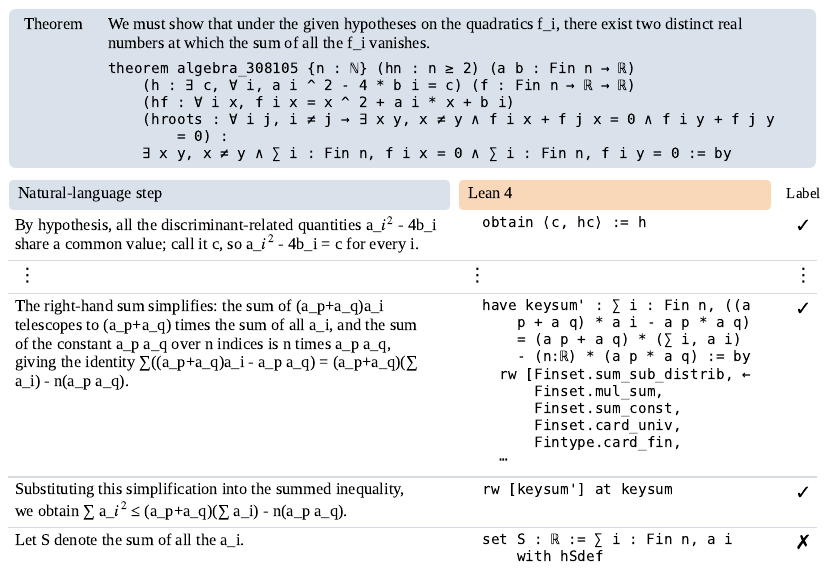}
\caption{An incomplete proof. The argument establishes the summed inequality and defines $S$ at step
19, then stops without producing the two distinct zeros. Under the convention of
Section~\ref{sec:pipeline} the final step of an argument that stops short is incorrect; every other
system accepts all nineteen steps, and LeanPRM-Verifier-4B$^{\dagger}$ flags the last.}
\label{fig:showcase_incomplete}
\end{figure}

\begin{figure}[t]
\centering
\includegraphics[width=\textwidth]{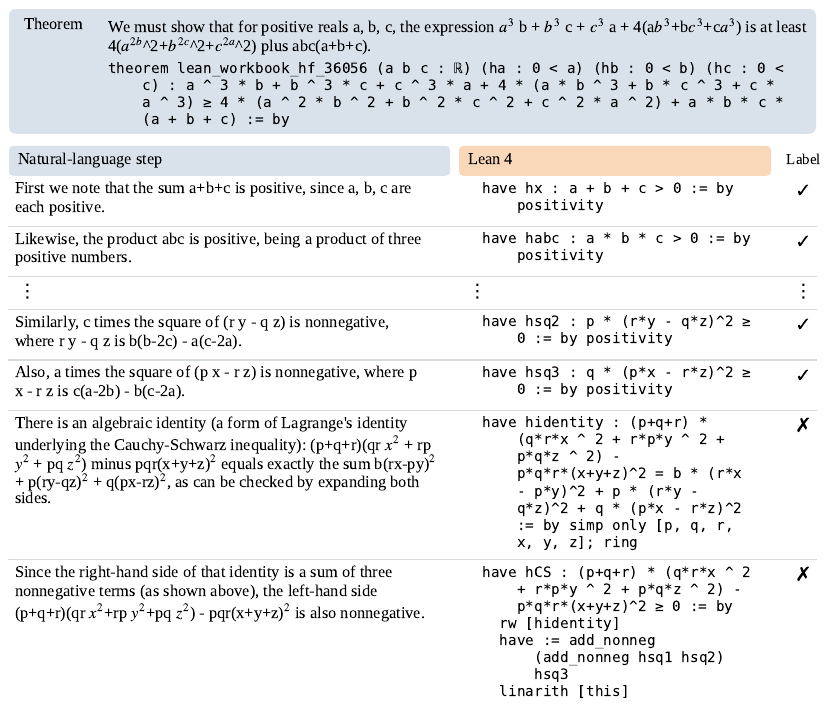}
\caption{The reverse case. Step 8 claims a Lagrange-type identity with the wrong coefficient on the
first square, so the checker's ring normalization fails and the class is algebraic identity error.
All four commercial systems and PRM800K-Verifier-4B flag step 8; LeanPRM-Verifier-4B$^{\dagger}$
accepts the proof.}
\label{fig:showcase_reverse}
\end{figure}

In the original reference run of each commercial system, 26 of the 200 proofs in
LeanPRM-Bench are mislabeled by all four systems.
LeanPRM-Verifier-4B$^{\dagger}$ labels 8 of them exactly, six incorrect proofs and two correct ones,
and over three decoding seeds it labels 29.5\% of the 26 exactly, against 7.7\% for the untrained
model and 11.5\% for PRM800K-Verifier-4B. The reverse case, all four commercial systems right and
the trained model wrong, covers 12 of the 128 proofs on which all four are right.
Figures~\ref{fig:showcase_pairing} and~\ref{fig:showcase_incomplete} show two of the eight and
Figure~\ref{fig:showcase_reverse} one of the twelve; the remaining cases follow the checker's
conventions, such as treating an argument that stops before its conclusion as incorrect at its last
step.

\section{Training details}
\label{app:training}

Given the theorem and its numbered natural-language steps, the verifier writes a reasoning section,
one label per step, and a verdict, and the verdict must be incorrect whenever any step is labeled
incorrect. Let $\mathrm{Rec}_C$ and $\mathrm{Rec}_I$ denote label recall on the steps that are truly
correct and truly incorrect. The step reward is $\mathrm{Rec}_C\,\mathrm{Rec}_I$ when the proof
contains an error and otherwise the indicator that every step is predicted correct; the task reward
is $0.7\,R_{\mathrm{step}} + 0.3\,R_{\mathrm{verdict}}$ with $R_{\mathrm{verdict}}$ the verdict
accuracy, and malformed output, a missing reasoning section, or an inconsistent verdict scores zero.
On an incorrect proof with both label classes, the product penalizes both the all-correct and the all-incorrect prediction.

All models are optimized with group-relative policy optimization, eight responses per prompt sampled
at temperature 1, batch size 64, learning rate $10^{-6}$, KL coefficient $10^{-3}$, and a
6{,}656-token response limit, on four H100 80GB GPUs, with validation and checkpoints every 50
updates. LeanPRM-Verifier-4B$^{\dagger}$ is a single run continued across three stages of data: 532
updates on an earlier pool of 3{,}449 training proofs produced by a joint generation procedure, 750
updates on the unfiltered pool of Section~\ref{sec:data} (its best validation checkpoint of 808),
and 300 updates on LeanPRM, 1{,}582 updates in total. Every LeanPRM verifier in Table~\ref{tab:matched}, and
PRM800K-Verifier-4B and NuminaMath-Solver-4B, trains from its base checkpoint for  \newrev{500 }updates. The
8B model uses the same configuration with parameter and optimizer offloading and rollouts
tensor-parallel across two GPUs.

\section{Evaluation details}
\label{app:eval}

The held-out split is scored by exact step-label accuracy, the fraction of proofs whose every label
is predicted correctly. Hard2Verify is scored by response-level balanced
accuracy over its 200 problems, ProcessBench by the mean F1 over its four subsets on all 3{,}400
test cases, and PRMBench by first-error accuracy on all 6{,}216 rows. On IMO-GradingBench the
four-way human grade is binarized as fully correct against everything else, candidates are split
into steps at paragraph boundaries with an even merge above thirty steps, and 13 of the 1{,}000
candidates exceed the context budget and are dropped, leaving 987. Every model receives the same
prompt layout and label parser. Hosted responses are normalized for Markdown formatting
before failed parses count as incorrect. We report mean and population standard deviation over
three runs, using seeds 42, 43, and 44 at temperature 1 and fixed checkpoints for open-weight models
and independent hosted runs without controlled seeds for commercial models. The Claude models are run through the Anthropic
API with extended thinking at high effort and a 16{,}000-token budget, retrying truncations
or refusals once at 32{,}000 tokens. Fable's system prompt omits the thinking-end tag. GPT-5.6
Sol is run through the Codex command-line interface on a ChatGPT subscription, one non-interactive
call per item at high reasoning effort with the shell, web search and every other tool disabled.
Codex prepends its own agent instructions to the paper's prompt and does not return the reasoning,
so this row is prompted through Codex rather than through the bare API; all 33{,}009
evaluations across its three runs completed without tool use.

\section{Model scale results}
\label{app:scale}

Table~\ref{tab:matched} gives the numbers behind Figure~\ref{fig:scaling}.

\begin{table}[h]
\centering
\small
\caption{Model scale at a fixed budget: LeanPRM verifiers trained from Qwen3 base models of three sizes for \newrev{500 }updates each with the same recipe, with the untrained models for comparison; mean $\pm$ population standard deviation over three decoding seeds. The 4B and 8B models are close on \newrev{LeanPRM-Bench}, \newrev{PRMBench }and \newrev{IMO-GradingBench, and }the 0.6B model is far behind.}
\label{tab:matched}
\setlength{\tabcolsep}{3.5pt}
\begin{tabular}{@{}lccccc@{}}
\toprule
& LeanPRM-Bench & Hard2Verify & ProcessBench & PRMBench & IMO-GB \\
& exact & response BA & avg.\ F1 & first err. & BA \\
\midrule
Qwen3-0.6B base & 22.7\,{\scriptsize$\pm$0.2} & 3.6\,{\scriptsize$\pm$1.7} & 3.2\,{\scriptsize$\pm$0.2} & 3.5\,{\scriptsize$\pm$0.5} & 48.4\,{\scriptsize$\pm$0.2} \\
Qwen3-4B (initial policy) & 48.2\,{\scriptsize$\pm$1.3} & 37.1\,{\scriptsize$\pm$4.5} & 74.3\,{\scriptsize$\pm$0.5} & 42.5\,{\scriptsize$\pm$0.5} & 53.8\,{\scriptsize$\pm$0.1} \\
Qwen3-8B base & 45.7\,{\scriptsize$\pm$1.9} & 52.4\,{\scriptsize$\pm$1.2} & 79.5\,{\scriptsize$\pm$0.3} & 47.6\,{\scriptsize$\pm$0.2} & 58.4\,{\scriptsize$\pm$0.6} \\
\midrule
LeanPRM-Verifier-0.6B & \newrev{54.0\,{\scriptsize$\pm$2.5} }& \newrev{61.8\,{\scriptsize$\pm$1.8} }& \newrev{34.9\,{\scriptsize$\pm$1.0} }& \newrev{27.7\,{\scriptsize$\pm$0.2} }& \newrev{58.4\,{\scriptsize$\pm$1.8} }\\
LeanPRM-Verifier-4B & \newrev{74.7\,{\scriptsize$\pm$1.8} }& \newrev{72.8\,{\scriptsize$\pm$0.9} }& \newrev{74.9\,{\scriptsize$\pm$0.4} }& \newrev{47.4\,{\scriptsize$\pm$0.2} }& \newrev{66.4\,{\scriptsize$\pm$1.1} }\\
LeanPRM-Verifier-8B & \newrev{75.0\,{\scriptsize$\pm$0.7} }& \newrev{66.7\,{\scriptsize$\pm$1.8} }& \newrev{79.3\,{\scriptsize$\pm$0.2} }& \newrev{48.0}\,{\scriptsize$\pm$0.3} & \newrev{67.2\,{\scriptsize$\pm$0.5} }\\
\bottomrule
\end{tabular}
\end{table}

\section{Labeling conventions and ProcessBench}
\label{sec:analysis}

We think most of the ProcessBench gap comes from labeling convention, but we have not tested this
directly. Our labels mark the first step that is not justified, while ProcessBench counts a missing
justification as an error only when it affects the validity of the solution, so on a true but
unjustified step our verifier should flag earlier than the reference. Across three checkpoints of
the run behind LeanPRM-Verifier-4B$^{\dagger}$, after 532, 1{,}282 and 1{,}332 updates, flags placed
strictly before the reference step rise from 158 to 236 to 379. Only the last two checkpoints share
training data, so the trend fits the explanation without isolating it; an audit of the early flags
would settle it.

\section{Further LeanPRM examples}
\label{app:examples}

Three further training examples in the format of Figure~\ref{fig:example}, chosen to show the three
kinds of attempt the pipeline produces.

\begin{figure}[h]
\centering
\includegraphics[width=\textwidth]{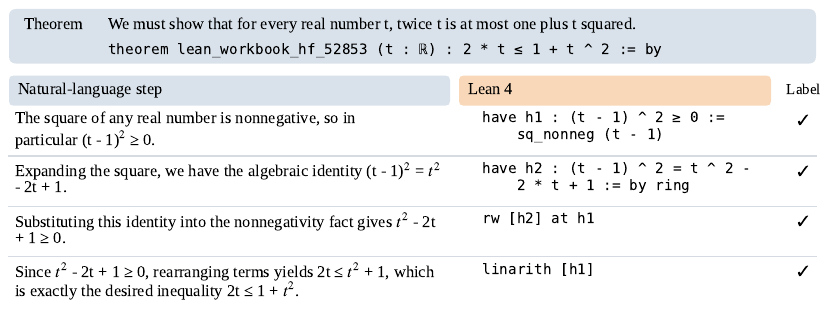}
\caption{A correct attempt: for real $t$, $2t \le 1 + t^2$. Every step elaborates, so every step is
labeled correct. The argument is the standard completion of the square, written as four claims each
realized by one tactic. Attempts of this kind account for 6{,}370 of the 10{,}634 theorems in
LeanPRM.}
\label{fig:example-correct}
\end{figure}

\begin{figure}[h]
\centering
\includegraphics[width=\textwidth]{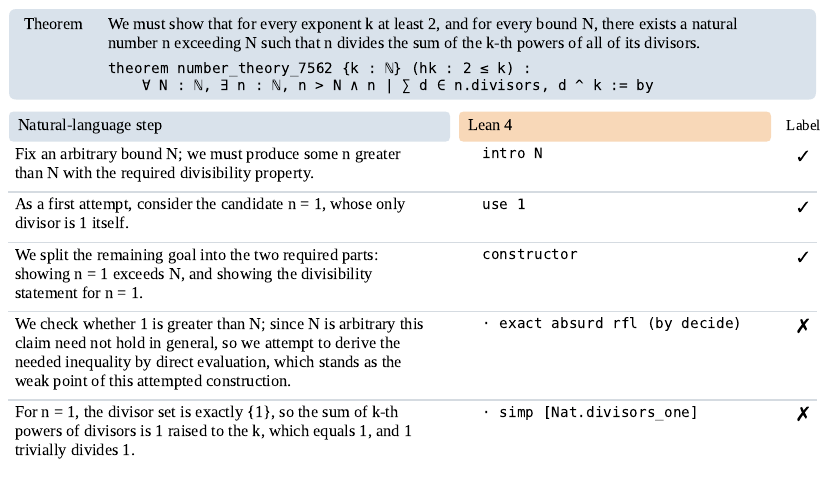}
\caption{A reasoning error whose text shows the slip. The statement asks, for $k \ge 2$ and every
bound $N$, for some $n > N$ dividing the sum of the $k$-th powers of its divisors. The attempt
proposes $n = 1$, which cannot exceed an arbitrary $N$, and step 4 concedes as much while trying to
discharge the bound anyway. Elaboration fails at that step; Stage 2 assigns the class invalid
inequality bound, and steps 4 and 5 are labeled incorrect.}
\label{fig:example-bound}
\end{figure}

\begin{figure}[h]
\centering
\includegraphics[width=\textwidth]{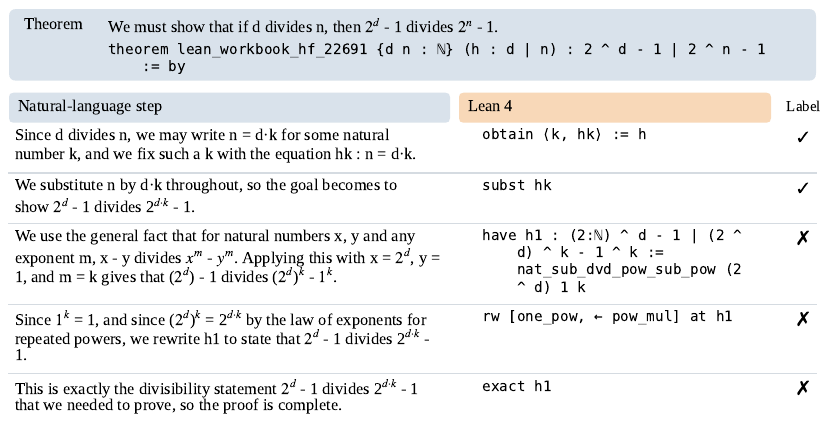}
\caption{An attempt from the labeled pool that is not part of LeanPRM. The statement is that $d \mid
n$ implies $2^d - 1 \mid 2^n - 1$. Step 3 invokes the true fact that $x - y$ divides $x^m - y^m$;
the mathematics is sound, but the Lean line names the library lemma in a form the pinned Mathlib
does not accept, so elaboration fails there and Stage 2 assigns theorem misapplication. The audit
judged steps of this class sound in 26 of 30 cases, which is why the formalization-side classes are
excluded from the corpus.}
\label{fig:example-formal}
\end{figure}

\section{Prompts}
\label{app:prompts}

The prompts below are reproduced from our code. Stage 1 receives the system prompt and a
user prompt that contains the theorem statement, the requirements listed, a worked example, and the
JSON schema of the expected output.

\begin{tcolorbox}[enhanced,breakable,colback=tracebg,colframe=traceframe,arc=3pt,boxrule=0.5pt,left=8pt,right=8pt,top=7pt,bottom=7pt]
\begin{Verbatim}[fontsize=\footnotesize,breaklines=true,breaksymbolleft={},fontfamily=fvm]
System prompt

You are an expert Lean 4 and mathematics proof writer.
Generate Lean proofs for dataset construction.  Each proof must be a sequence
of small tactic steps, and every step must have a matching mathematical natural
language explanation that does not mention Lean syntax, tactic names, formal
verification, or code.

The Lean code and the natural-language proof must be faithful to each other:
the Lean proof should implement the same mathematical argument described in the
natural language, not skip over several natural-language claims with one opaque
automation call.

Do not ban arithmetic discharge tactics outright, but do not use `linarith`,
`nlinarith`, or `omega` as opaque one-shot proofs of the main argument.  If one
of these tactics is used, the preceding steps must expose the concrete
mathematical facts, identities, cases, or inequalities it relies on.  Never use
`sorry`, `admit`, or any placeholder proof term.

Two hard requirements for the natural language:

1. Each NL description must be 1:1 with its corresponding Lean step --
   describing exactly what that step does mathematically, not a higher-level
   summary and not a vague gloss.
2. Concatenating all NL descriptions in order must read as a complete,
   standalone mathematical proof that a student could have written and a
   mathematician could follow without ever seeing the Lean code.  It must not
   read like comments about code or a list of tactic explanations.

User prompt

Generate one independent proof attempt for this Lean theorem:

```lean
{lean_statement}
```

Requirements:
- Use Lean 4 with Mathlib.
- Keep the theorem name and statement exactly as given.
- Do not use `linarith`, `nlinarith`, or `omega` as opaque one-shot proofs.
  They are allowed only as final arithmetic discharge after the preceding steps
  have made the algebraic situation, case split, divisibility fact, modular
  computation, or inequality combination explicit enough that the NL remains
  1:1 and mathematically informative.
- Do not use `sorry`, `admit`, `by_contra!` as a placeholder, or any unfinished
  proof placeholder.
- Do not use one opaque tactic to hide the main argument.  If a proof needs
  algebra, inequalities, case splits, contradiction, divisibility facts, or
  modular arithmetic, introduce those facts as explicit intermediate `have`,
  `calc`, `rw`, or case-analysis steps before any final arithmetic discharge.
- The proof must be step-by-step: step 0 is the theorem line ending in `:= by`,
  and each later step is one Lean tactic or one tightly scoped Lean proof line.
- Keep the proof concise. Aim for at most 40 steps and do not split routine
  rewriting or arithmetic into dozens of tiny steps.
- Each step must include a natural-language mathematical explanation matching
  that exact Lean step 1:1 (see rules and example below).
- Natural language must explain the mathematics only.  Do not mention Lean,
  tactics, automation, syntax, or the compiler.
- The natural language must not describe a stronger or more detailed argument
  than the Lean line actually proves.  If the NL says "we write k = 4q + 1",
  the Lean line must actually establish that statement in the corresponding
  case; if the NL cites an algebraic identity, the Lean line must establish that
  identity or use it directly.
- It is acceptable if the proof attempt is mathematically wrong, but avoid
  syntax errors and malformed Lean where possible.
- Return only JSON.  No Markdown fences.

[the natural-language rules, a worked example, and the JSON schema follow]

\end{Verbatim}

\end{tcolorbox}

Stage 2 receives the following instructions together with the ten error classes of
Appendix~\ref{app:classes}, the Lean statement, the compiler output, the frozen steps, and a JSON
schema for its verdict.

\begin{tcolorbox}[enhanced,breakable,colback=tracebg,colframe=traceframe,arc=3pt,boxrule=0.5pt,left=8pt,right=8pt,top=7pt,bottom=7pt]
\begin{Verbatim}[fontsize=\footnotesize,breaklines=true,breaksymbolleft={},fontfamily=fvm]
Classify this immutable Stage-1 Lean/NL proof attempt after Lean verification and, only when appropriate, return minimal surface-form Lean syntax patches.

Stage 1 is the true blind attempt. Its natural-language descriptions and
mathematical plan are frozen. You MUST NOT rewrite, improve, complete, or correct
them. Compiler feedback is available only for classification and private syntax
repair; it must never leak into the natural-language trajectory.

This is the second and final LLM call for this datapoint. Decide one verdict:

- "correct": Lean verification already SUCCEEDED and the natural-language proof is
  faithful to the Lean proof.
- "reasoning_error": Lean verification FAILED because the mathematical proof
  attempt is false, unjustified, or incomplete at the first failing step. This
  includes a missing branch, missing direction, unclosed goal, or an argument
  that stops before proving the theorem. These are valuable natural solver
  failures and MUST be retained as reasoning errors, not discarded. If the
  attempt simply stops with an unclosed goal, use its final provided step as
  `first_error_step`, not step 0.
- "repair": Lean verification FAILED for a purely MECHANICAL/code reason while the
  underlying mathematics is VALID, such as punctuation, parentheses, namespace,
  a renamed identifier with exactly the same proposition, or an explicit type or
  coercion annotation expressing the same term. Return only the changed Lean lines.
- "discard": the statement is ill-posed, the output is malformed, or the NL and
  Lean silently describe different mathematics. Do NOT discard merely because
  the proof is incomplete or would need substantial new mathematics to finish.

REPAIR RULES (only when verdict is "repair"):
- Fix ONLY surface-form Lean syntax or elaboration. Do NOT change the mathematics.
- Do NOT alter step 0 in any way.
- Do NOT weaken the theorem, and do NOT use `sorry`, `admit`, `native_decide`, or any placeholder/cheat tactic.
- Do NOT add, remove, merge, split, or reorder steps.
- Do NOT change a proposition, expression, witness, assumption, case split,
  intermediate claim, tactic kind, tactic strategy, or tactic arguments that carry
  mathematical content.
- Do NOT swap tactics or add hints, facts, automation, or "small nudges."
- If it cannot compile without adding or changing substantive mathematics, use
  "reasoning_error" when the attempt is incomplete or mathematically invalid.

For any verdict other than "repair", return no syntax patches.

[the list of ten error classes, the Lean statement, the compiler output, the steps, and the JSON schema follow]

\end{Verbatim}

\end{tcolorbox}

The verifier is trained and evaluated with the prompt below; every benchmark is rendered into the
same layout, and the parser reads the label lines after the thinking section.

\begin{tcolorbox}[enhanced,breakable,colback=tracebg,colframe=traceframe,arc=3pt,boxrule=0.5pt,left=8pt,right=8pt,top=7pt,bottom=7pt]
\begin{Verbatim}[fontsize=\footnotesize,breaklines=true,breaksymbolleft={},fontfamily=fvm]
System prompt

You are a mathematical proof verifier. Reason carefully about every numbered natural-language step. Once a step is incorrect, every later step is incorrect because it depends on an invalid proof state. Use your thinking section for the analysis. After </think>, return only the requested labels with no explanation.

User prompt

Problem:
<problem statement>

Solution:
<step1>...</step1>
<step2>...</step2>
...

Label all 11 steps. Then give the overall proof verdict.
Use exactly this format:
Step 1: correct or incorrect
Step 2: correct or incorrect
...
Overall: correct or incorrect

\end{Verbatim}

\end{tcolorbox}

The proof-level audit of Section~\ref{sec:audit} uses the first prompt below and the per-class audit
the second; in both the judge never sees our labels.

\begin{tcolorbox}[enhanced,breakable,colback=tracebg,colframe=traceframe,arc=3pt,boxrule=0.5pt,left=8pt,right=8pt,top=7pt,bottom=7pt]
\begin{Verbatim}[fontsize=\footnotesize,breaklines=true,breaksymbolleft={},fontfamily=fvm]
You are auditing a dataset of natural-language mathematical proofs with step-level correctness labels.
You will see a problem statement and a proof split into numbered steps. Do NOT assume the proof is correct or incorrect.

Tasks:
1. Read every step. Decide the index of the FIRST step that is mathematically incorrect or unjustified (a false claim, an invalid inference, a missing case, or a step that does not follow from what precedes it). If every step is valid and the proof is complete, answer 0.
2. If you found an error, rate how natural it is on a 1-5 scale (5 = the kind of mistake a strong student would plausibly make; 1 = obviously artificial, e.g. a random symbol change, a contradiction of the previous line, or a nonsensical claim).
3. Say whether the text leaks formalization artefacts (mentions of Lean, Mathlib, tactics, `theorem`/`lemma` declarations, sorry, type annotations) that a natural proof would not contain: yes or no.
4. One sentence explaining the first error, or "none".

Answer with exactly this JSON object and nothing else:
{"first_incorrect_step": <int>, "naturalness": <int or null>, "leak": "yes"|"no", "explanation": "<one sentence>"}

\end{Verbatim}

\end{tcolorbox}

\begin{tcolorbox}[enhanced,breakable,colback=tracebg,colframe=traceframe,arc=3pt,boxrule=0.5pt,left=8pt,right=8pt,top=7pt,bottom=7pt]
\begin{Verbatim}[fontsize=\footnotesize,breaklines=true,breaksymbolleft={},fontfamily=fvm]
You are auditing step-level labels on natural-language mathematical proofs. A proof checker flagged ONE step as the first incorrect step. Judge that step on its mathematical merits only.

You will see the problem, the steps before the flagged step (assume they are correct), and the flagged step.

Answer:
1. valid: is the flagged step's claim mathematically true AND adequately justified by what precedes it (a strong grader would accept it)? "yes" or "no".
2. If "no", naturalness of the mistake on a 1-5 scale (5 = a plausible student mistake, 1 = artificial or nonsensical); null if "yes".
3. One sentence: what is wrong, or why it is fine.

Answer with exactly this JSON object and nothing else:
{"valid": "yes"|"no", "naturalness": <int or null>, "explanation": "<one sentence>"}

\end{Verbatim}

\end{tcolorbox}

\section{Hyperparameters}
\label{app:hparams}

Table~\ref{tab:hparams} lists the training configuration. Every trained model in the paper, the
baselines included, uses it; only the training data and, for the scaling runs, the base model
differ, and the 8B runs also use offloading and tensor parallelism to fit in memory.

\begin{table}[h]
\centering
\footnotesize
\caption{Training configuration shared by every trained model.}
\label{tab:hparams}
\begin{tabular}{ll}
\toprule
Base models & Qwen3-0.6B, Qwen3-4B, Qwen3-8B, thinking mode enabled \\
Optimizer & GRPO as implemented in VERL, learning rate $10^{-6}$ \\
Batch & 64 prompts per update, 8 responses per prompt \\
Clipping & ratio clipped at 0.20 below and 0.26 above, dual-clip constant 10 \\
Regularization & KL coefficient $10^{-3}$, low-variance estimator, no entropy bonus \\
Loss aggregation & sequence mean of token means \\
Sequence lengths & 8{,}192 prompt tokens, 6{,}656 response tokens \\
Rollouts & vLLM at temperature 1; validation greedy, one response per prompt \\
Schedule & checkpoints and validation every 50 updates \\
Hardware & four H100 80GB GPUs \\
 & 8B model with parameter and optimizer offloading \\
 & and rollouts tensor-parallel over two GPUs \\
\bottomrule
\end{tabular}
\end{table}

\end{document}